%% file: lrec2026-example.tex
\documentclass[10pt, a4paper]{article}

\usepackage[final]{lrec2026} 

\usepackage{amssymb}
\usepackage{amsmath}

\usepackage{booktabs}   
\usepackage{array}  
\usepackage{microtype}  
\usepackage{xcolor}
\usepackage{multirow}
\usepackage{makecell}
\usepackage{subcaption}
\usepackage{url}
\usepackage{rotating}
\usepackage{enumitem} 

\title{Zero-Shot to Full-Resource: Cross-lingual Transfer Strategies for Aspect-Based Sentiment Analysis}

\name{Jakob Fehle$^1$, Nils Constantin Hellwig$^1$, Udo Kruschwitz$^2$, Christian Wolff$^1$}

\address{
$^1$Media Informatics Group, University of Regensburg, Regensburg, Germany \\
$^2$Information Science Group, University of Regensburg, Regensburg, Germany \\
jakob.fehle@ur.de, nils-constantin.hellwig@ur.de, udo.kruschwitz@ur.de, christian.wolff@ur.de
}

\abstract{
Aspect-based Sentiment Analysis (ABSA) extracts fine-grained opinions toward specific aspects within text but remains largely English-focused despite major advances in transformer-based and instruction-tuned models.
This work presents a multilingual evaluation of state-of-the-art ABSA approaches across seven languages (English, German, French, Dutch, Russian, Spanish, and Czech) and four subtasks (ACD, ACSA, TASD, ASQP). We systematically compare different transformer architectures under zero-resource, data-only, and full-resource settings, using cross-lingual transfer, code-switching and machine translation.
Fine-tuned Large Language Models (LLMs) achieve the highest overall scores, particularly in complex generative tasks, while few-shot counterparts approach this performance in simpler setups, where smaller encoder models also remain competitive. Cross-lingual training on multiple non-target languages yields the strongest transfer for fine-tuned LLMs, while smaller encoder or seq-to-seq models benefit most from code-switching, highlighting architecture-specific strategies for multilingual ABSA.
We further contribute two new German datasets, an adapted GERestaurant and the first German ASQP dataset (GERest), to encourage multilingual ABSA research beyond English.
 \\ \newline \Keywords{Aspect-based Sentiment Analysis, Cross-Lingual, Resources, Large Language Models} }

\begin{document}

\maketitleabstract

\section{Introduction}

\input{sections/introduction}
\section{Related Work}
\input{sections/related_work}
\section{Methodology}
\input{sections/methodology}
\section{Results}
\input{sections/results}
\section{Conclusion \& Future Work}
\input{sections/conclusion}
\section*{Limitations \& Ethical Considerations}

\input{sections/limitations}


\section{Bibliographical References}\label{sec:reference}

\bibliographystyle{lrec2026-natbib}
\bibliography{lrec2026-example}

\input{sections/appendix}

\end{document}

%% file: sections/introduction.tex
Aspect-based Sentiment Analysis (ABSA) has become a central task for mining fine-grained opinions, aiming to detect sentiment toward specific aspects within text. Despite substantial methodological advances, from transfer learning-based classifiers \cite{Cai2020-km, Cui2024-vx} to instruction-tuned large language models (LLMs) \cite{Scaria2024-wc, Smid2024-ms}, research has remained largely English-focussed, driven by the abundance of English benchmark datasets \cite{Chebolu2023-li}.


\citet{Smid2025-el} emphasize that ABSA generalization to non-English languages remains challenging due to limited multilingual data diversity, high sensitivity to translation quality, and structural divergence between languages, and they further emphasize that cross-lingual transfer remains inconsistent across languages and tasks. Cross-lingual studies \citep{Lin2023-bq, Zhang2025-ih, Smid2025-kw} similarly report inconsistent transfer, with multilingual encoders such as mBERT or XLM-R showing substantial performance gaps across languages even under comparable supervision. While multilingual pre-trained language models such as mBERT \cite{Devlin2019-gx} and mT5 \cite{Xue2021-pr}, or more recent LLMs including GPT-4 \cite{openai2024gpt4technicalreport}, LLaMA 3 \cite{Dubey2024-x}, and Gemma 3 \cite{gemmateam2025gemma3technicalreport}, have substantially improved zero-shot and cross-lingual transfer capabilities, systematic evaluations across diverse ABSA subtasks and multiple languages remain limited, particularly for complex generative tasks beyond End-to-End ABSA (E2E) or single-element extraction \cite{Smid2025-el}.

We address this gap by conducting a comprehensive evaluation of state-of-the-art (SOTA) methods across four established ABSA sub-tasks: Aspect Category Detection (ACD), Aspect Category Sentiment Analysis (ACSA), Target Aspect Sentiment Detection (TASD), and Aspect Sentiment Quad Prediction (ASQP). To ensure cross-lingual comparability, we base our experiments on the SemEval-2016 restaurant datasets \cite{Pontiki2016-hj}, the most widely adopted multilingual ABSA resource \cite{Hua2024-vs}, covering six languages (English, French, Spanish, Dutch, Russian, Turkish). We extend this benchmark with German \cite{Hellwig2024-wi} and Czech \cite{Smid2024-hp} datasets following the same schema, and \textbf{contribute the first German ASQP dataset}, \textit{GERest}, to enable cross-lingual ASQP evaluation.

We systematically compare three modeling paradigms: 
\begin{enumerate}[label=(\alph*)]
    \item \textbf{Encoder-only classification}, which conceptualize ABSA as a supervised multi-label classification problem, including BERT-based architectures and graph convolutional networks (GCNs). 
    \item \textbf{Sequence-to-sequence text generation}, which reformulates ABSA as a structured text generation task using T5-based models with predefined templates for input and output.
    \item \textbf{Decoder-only LLMs}, encompassing both few-shot in-context prompting and instruction fine-tuning for structured sentiment extraction.
\end{enumerate}

This setup allows us to compare how different architectures and learning paradigms handle cross- and multilingual conditions under a unified experimental framework. Experiments are conducted under three resource conditions: 
\begin{enumerate}
    \item In the \textbf{zero-resource setting}, neither annotated data nor language-specific models are available; models must rely solely on cross-lingual transfer capabilities.
    \item In the \textbf{data-only setting}, annotated training data in the target language is available, but no dedicated language-specific model exists, requiring multilingual models to adapt to the language.
    \item In the \textbf{full-resource setting}, both annotated data and language-specific pre-trained models are available, allowing us to assess the performance ceiling for each language.
\end{enumerate}

To enhance zero-resource settings, we apply code-switching and machine-translation augmentation \cite{Zhang2021-qu} to generate pseudo-training data from English.

Our study provides a comprehensive empirical analysis of multilingual ABSA. We quantify the trade-offs between multilingual and language-specific models, assess the limits of transfer-based methods in low-resource conditions, and highlight practical implications for adapting SOTA ABSA techniques across linguistically diverse settings. All code and results are available at GitHub.\footnote{GitHub: \url{https://github.com/JakobFehle/Cross-lingual-Transfer-Strategies-for-ABSA}}

%% file: sections/related_work.tex

In recent years, ABSA has seen substantial progress through both classification-based and generative modeling approaches. 

\subsection{State-of-the-Art Modeling Approaches for ABSA}

Recent advances in ABSA span a continuum from supervised classification to generative and instruction-based approaches. For simpler subtasks such as ACD and ACSA, transformer-based classifiers like BERT-CLF~\cite{Fehle2023-zj, Hellwig2024-wi}, graph-based models such as Hier-GCN~\cite{Cai2020-km}, and attention-augmented variants like ECAN~\cite{Cui2024-vx} have established strong baselines. More complex tasks, including TASD and ASQP, are increasingly modeled as a text generation problem using sequence-to-sequence architectures such as T5~\cite{Raffel2020-ky}, where predefined~\cite{Zhang2021-qu} or dynamically ordered templates~\cite{Hu2022-dx, Gou2023-hx} are used to structure the outputs.

LLMs have further advanced ABSA in both supervised and unsupervised settings. Instruction fine-tuning, which reformulates ABSA inputs as natural language prompts, achieves SOTA results across multiple benchmarks~\cite{Varia2023-lc, Simmering2023-x, Smid2024-ms, Fehle2026-ws}. These studies show that instruction-tuned models like T5 or LLaMA outperform traditional fine-tuning and few-shot prompting, with prompt design playing a minor role once models are instruction-aligned~\cite{Simmering2023-x}. \citet{Smid2024-ms} further demonstrate that fine-tuned open-source LLMs can outperform proprietary ones, though most of current evaluations remain limited to English.


Recent research also focuses on ABSA under low-resource conditions, which has shifted from manual corpus creation~\cite{Akhtar2016-gi, Rani2020-tg} to generative and hybrid annotation methods~\cite{Hu2022-dx, Gou2023-hx, Hellwig2025-ss} that leverage template variation and synthetic data for improved robustness.


Overall, ABSA has evolved from static classification toward flexible generative and instruction-based modeling, with augmentation methods providing the groundwork for multilingual and cross-lingual evaluation.

\subsection{Cross-Lingual ABSA}

Cross-lingual ABSA research has been driven by the availability of multilingual benchmarks, most notably the SemEval-2016 Task 5 datasets~\cite{Pontiki2016-hj}, which provide manually annotated restaurant, hotel, and laptop reviews across eight languages under a unified schema. Their accessibility has made them the de facto standard for multilingual evaluation. Later resources such as MultiAspectEmo~\cite{Szolomicka2022-il} and M-ABSA~\cite{Wu2025-x} expanded language coverage through translation or alignment but often lack gold-standard quality.

Early cross-lingual ABSA approaches transferred knowledge via bilingual embeddings or machine translation~\cite{Barnes2016-kk, Jebbara2019-pc, Garcia-Pablos2018-oe}, but transfer performance remained limited for low-resource languages. With the advent of multilingual pre-trained encoders such as mBERT and XLM-R, encoder-only architectures became the dominant paradigm~\cite{Phan2021-gc, Van-Thin2023-eb}. Later studies enhanced zero-shot transfer through augmentation techniques like aspect code-switching~\cite{Zhang2021-kz}, contrastive alignment and distillation~\cite{Lin2023-bq}, and synthetic data generation or consistency regularization~\cite{Wu2025-xddd, Smid2025-xz}, showing that unlabeled or LLM-generated target-language data can partially compensate for missing annotations.



\citet{Smid2025-el} provide a comprehensive survey of cross-lingual ABSA and identify remaining challenges, including reliance on translation quality, limited language diversity of available benchmarks, and weak generalization of models to morphologically rich languages. They further highlight the need for unified cross-lingual evaluations across multiple ABSA subtasks. 

Recent work extends this direction with prompting- and generation-based methods: constrained decoding with mT5 and LLaMA-3 improves zero-shot transfer~\cite{Smid2025-kw}, while few in-language examples can already yield substantial gains~\cite{Smid2025-mr}, which underlines the strong potential of minimal in-language supervision as a cost-efficient alternative to large-scale annotation. Similarly, \citet{Wu2024-bz} report that GPT- and LLaMA-based prompting captures sentiment cross-lingually but still trails fine-tuned models.

Overall, the literature reveals three main trends: (1) encoder-only models remain competitive for simpler classification-level tasks; (2) constrained or template-based generation performs best on complex subtasks such as TASD or ASQP; and (3) decoder-only LLMs generalize well but typically require adaptation or fine-tuning. Building on these insights, our work systematically compares all three paradigms, classification, sequence generation, and LLM-based methods, across multiple languages. In contrast to prior studies that rely on individual pretrained models or isolated techniques, we explore combinations of code-switching, machine translation, and multilingual training, in combination with several SOTA approaches.

%% file: sections/methodology.tex
\subsection{Tasks}

ABSA comprises a number of subtasks that differ in the level of detail and the type of information they extract from the text. In this work, we focus on four common ABSA tasks with different levels of granularity that are supported by the structure and annotations of our multilingual datasets: Aspect Category Detection (ACD), Aspect Category Sentiment Analysis (ACSA), Target Aspect Sentiment Detection (TASD), and Aspect Sentiment Quad Prediction (ASQP) ~\cite{Zhang2023-wv}. See Table~\ref{tab:absa-tasks} for examples illustrating the input and expected output of each task:

\begin{table}[t]
\centering
\includegraphics[width=\linewidth]{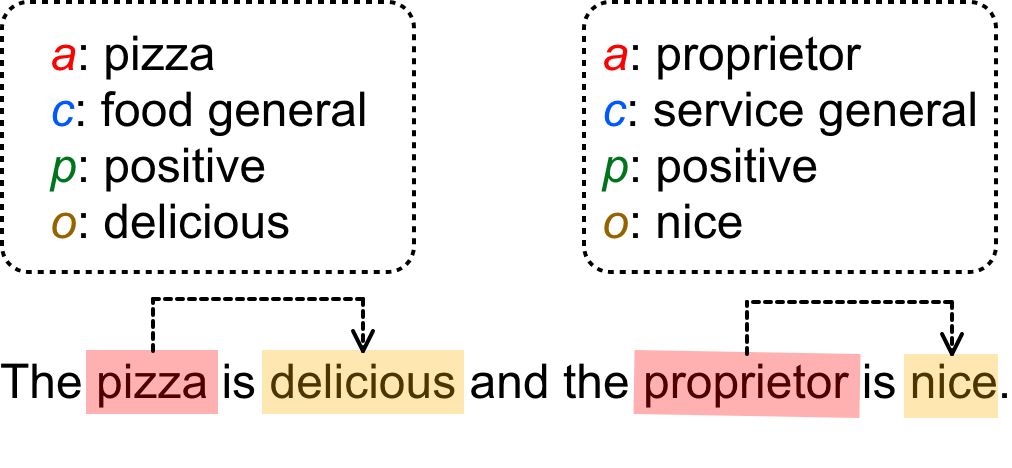}
\resizebox{\columnwidth}{!}{
\begin{tabular}{@{}lc@{}}
\toprule
\textbf{Subtask} & \textbf{Output} \\
\midrule
Aspect Category Detection (ACD) & $(\textcolor[HTML]{0058FF}{c})$ \\
Aspect Category Sentiment Classification (ACSA) & $(\textcolor[HTML]{0058FF}{c}, \textcolor[HTML]{00741B}{p})$ \\
Target Aspect Sentiment Detection (TASD) & $(\textcolor[HTML]{FF0000}{a}, \textcolor[HTML]{0058FF}{c}, \textcolor[HTML]{00741B}{p})$ \\
Aspect Sentiment Quad Prediction (ASQP) & $(\textcolor[HTML]{FF0000}{a}, \textcolor[HTML]{0058FF}{c}, \textcolor[HTML]{966200}{o}, \textcolor[HTML]{00741B}{p})$ \\
\bottomrule
\end{tabular}
}
\caption{Overview of ABSA subtasks used in this study with their expected outputs. Input for all tasks is the text sentence.}
\label{tab:absa-tasks}
\end{table}

\begin{itemize}
    \item ACD focuses on identifying all aspect categories mentioned or implied in a given input text.
    \item ACSA adds sentiment polarity classification (positive, neutral, negative) for each detected aspect.
    \item TASD further builds upon ACSA by requiring the detection of the exact text spans that represents the target of the expressed sentiment (aspect term). For implicit aspect terms, i.e., if no explicit phrase represents the aspect category in the text, the model is expected to return the value "\texttt{NULL}".
    \item ASQP aims to jointly extract all four components of a sentiment expression: the aspect term, its corresponding opinion term, the associated aspect category, and the expressed sentiment polarity, thus unifying extraction and sentiment classification in a single structured task.
\end{itemize}

\subsection{Datasets}
To evaluate the multilingual applicability of SOTA methods for ABSA, we rely on a diverse set of datasets covering multiple languages and ABSA subtasks.

\subsubsection{ACD, ACSA, and TASD}

For ACD, ACSA, and TASD, we use the multilingual restaurant review datasets from SemEval-2016 Task 5~\cite{Pontiki2016-hj}, employing the standardized training and test splits for five languages (English, Spanish, French, Dutch, and Russian) annotated with aspect terms, aspect categories, and sentiment polarity. All datasets share a unified 12-category schema, ensuring consistency across languages.

To expand the linguistic coverage, we added two further corpora: (1) GERestaurant~\cite{Hellwig2024-wi}, a German dataset we aligned to the SemEval schema through manual re-annotation of aspect categories, and (2) the Czech dataset by \citet{Smid2024-hp}, which already adheres to the same guidelines. This extended collection allows evaluation of cross-lingual robustness beyond the original SemEval languages.



\subsubsection{ASQP}
Since the original SemEval 2016 Task 5 datasets do not contain annotations for explicit opinion phrases, they are not directly suitable for ASQP. 



For English, we adopt the dataset by \citet{Zhang2021-qu}, which extends SemEval-2016 with opinion-phrase labels and serves as the standard benchmark for generative ASQP models~\cite{Gou2023-hx, Bai2024-dq}.

To extend the ASQP evaluation beyond English and enable a controlled cross-lingual comparison, we created a German counterpart (\textit{GERest}). Starting from the existing GERestaurant dataset~\cite{Hellwig2024-wi}, we manually annotated opinion terms for a subset matching the size of the English ASQP-Rest16 corpus (1,264 train; 316 dev; 544 test samples), following the same annotation guidelines. 

Further details on the dataset and its annotation process are provided in Appendix~\ref{appendix:gerest}. To prevent potential contamination of the GERest training and test sets in future language models, the dataset is not publicly released but is available upon reasonable request from the authors.


\subsection{Preprocessing}

\begin{figure*}[ht]
    \centering
    \includegraphics[width=\textwidth]{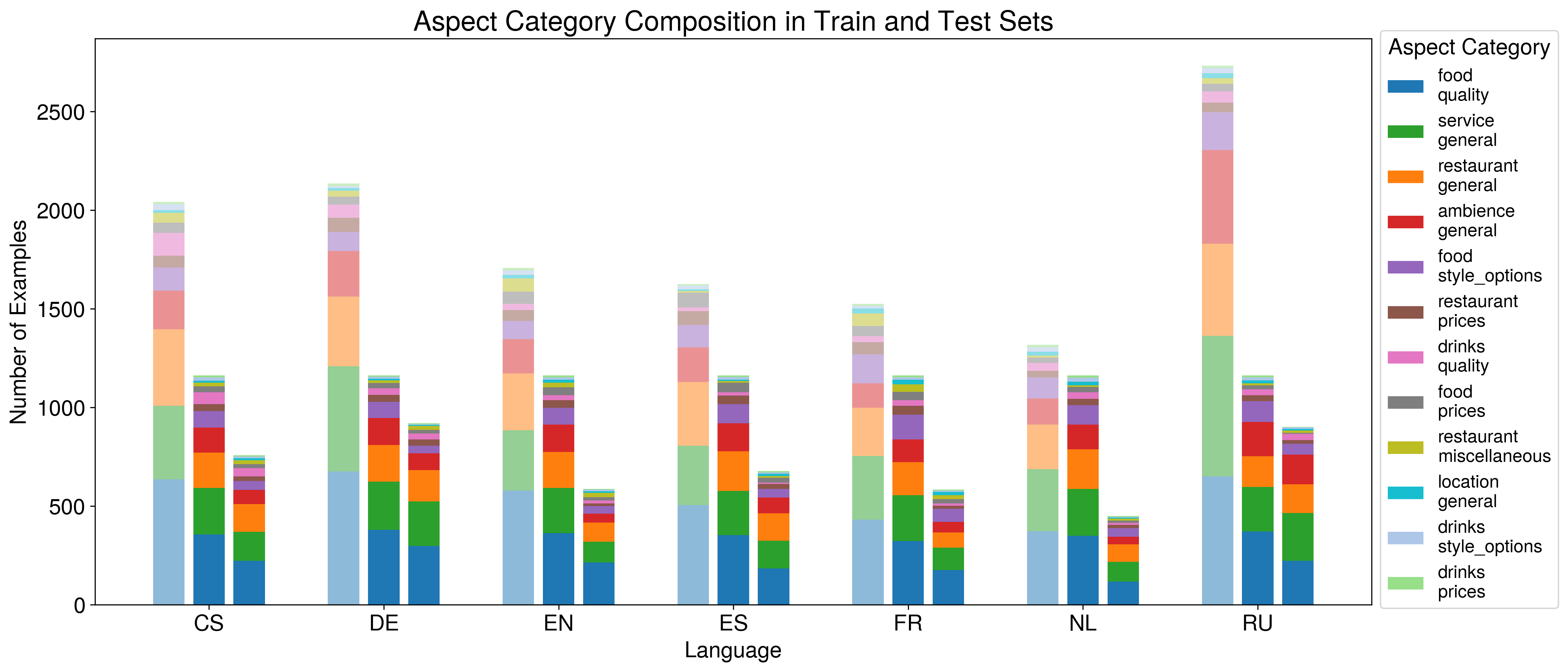}
    \caption{
        The diagram illustrates both the absolute dataset sizes and aspect category distributions across languages and splits. For each language, the three grouped bars represent (from left to right): original train, balanced train, and original test set. Each bar is stacked by relative aspect category distribution. Lighter colors indicate the original train split, while darker bars represent the balanced train and original test splits used in our experiments.
    }
    
    \label{fig:aspect_distribution_train}
\end{figure*}

\label{sec:preprocessing}
To ensure that cross-lingual comparisons are not affected by differences in class distribution, we constructed balanced subsets for each language that share a similar distribution across the 12 predefined aspect categories. Instead of preserving the original imbalanced distributions, we aimed to approximate a unified target distribution applicable to all datasets.

We first computed the global frequency of each category across languages and determined the maximum common support per class, which is the largest number of samples consistently available in all datasets. These values served as sampling targets, allowing for minor flexibility. We then randomly sampled instances per language to match this shared distribution as closely as possible, resulting in seven balanced training sets of 1,162 examples each. Figure~\ref{fig:aspect_distribution_train} illustrates the resulting category alignment and dataset sizes across all languages.



\subsection{Evaluated Approaches}

To ensure comparability across languages, we implement representative SOTA ABSA methods originally developed for the English SemEval-2016 restaurant dataset. Since most of these methods have not been systematically applied to other languages, we adopt their original hyperparameters and configurations with minimal adjustments. Methods were selected based on reported performance and the availability of reproducible implementations. 



To assess the cross‐lingual portability of SOTA ABSA methods, we experiment with three complementary paradigms: (1) encoder-based classification, (2) sequence-to-sequence generation, and (3) decoder-only LLMs with prompting or instruction fine-tuning.

\subsubsection{Encoder-Only Classification}

\begin{itemize}
  \item \textbf{BERT-CLF:} A multi-label transformer classifier predicting aspect categories (e.g., \texttt{food}) or aspect-sentiment pairs (e.g., \texttt{food:positive}) per sentence, following \citet{Fehle2023-zj} and \citet{Hellwig2024-wi}.
  \item \textbf{Hier-GCN:} Extends BERT with hierarchical graph convolutional layers to capture dependencies between aspects and sentiments \cite{Cai2020-km}.
\end{itemize}
\paragraph{Models:} Multilingual \textit{mBERT}\footnote{\url{https://huggingface.co/google-bert/bert-base-multilingual-cased}} \cite{Devlin2019-gx} and language-specific variants, such as \textit{ruBERT}\footnote{\url{https://huggingface.co/DeepPavlov/rubert-base-cased}} \cite{Kuratov2019-x} for Russian.  

\subsubsection{Seq-2-Seq Text Generation}
\begin{itemize}
  \item \textbf{DLO:} Dynamic Label Ordering \cite{Hu2022-dx} reformulates ABSA as a generative task by dynamically augmenting and reordering output tuples (e.g., for TASD, ASQP), improving alignment between input and structured outputs.
\end{itemize}
\paragraph{Models:} Multilingual \textit{mT5}\footnote{\url{https://huggingface.co/google/mt5-base}}~\cite{Xue2021-pr} and, where available, monolingual T5 variants, e.g., \textit{ruT5}\footnote{\url{ai-forever/ruT5-base}}~\cite{Zmitrovich2023-qz}.  

\subsubsection{Decoder-Only LLMs}
\begin{itemize}
  \item \textbf{Zero-/Few-Shot Prompting:} This approach uses instruction- and few-shot-based in-context learning (ICL) on a LLM with either zero or a predefined set of annotated ABSA examples embedded in prompts. 
  \item \textbf{Instruction Fine-Tuning:} Supervised fine-tuning of LLMs on task-specific instructions, updating model weights for ABSA generation. We use the same prompt structure as in the zero-/few-shot setting to maintain consistency across methods. 
\end{itemize}
\paragraph{Models and Implementation:} We use the same multilingual checkpoints for all languages: Gemma 3 27B\footnote{\url{https://huggingface.co/google/gemma-3-27b-it}} \cite{gemmateam2025gemma3technicalreport} with up to 50 annotated ABSA examples for zero-/few-shot prompting and LLaMA 3.1 8B\footnote{\url{https://huggingface.co/meta-llama/Llama-3.1-8B}} \cite{Dubey2024-x} for instruction fine-tuning. The selection of LLMs and amount of few-shots is based on results achieved by \citet{Hellwig2025-sd}. 
Prompt templates are based on \citet{Gou2023-hx}, adapted to the specific subtasks and translated into each target language to ensure structural and linguistic consistency. Prompt examples for English are provided in Appendix~\ref{app:prompt-examples}. 

Fine-tuning utilizes Parameter-Efficient Fine-Tuning (PEFT) \cite{peft} and is performed with Quantized Low-Rank Adaption (QLoRA) \cite{Dettmers2023-sv} using the \textit{unsloth}\footnote{\url{https://github.com/unslothai/unsloth}} framework.


\subsection{Systematic Evaluation and Hyperparameter Calibration}
To ensure fair cross‐lingual comparisons, we adopt a two‐stage evaluation strategy for each model and subtask. First, we determine the optimal number of training epochs via an 80/20 split of the source training data, keeping all other hyperparameters (learning rate, batch size, optimizer settings) constant across languages and tasks. We explore predefined epoch ranges depending on model type: 20~–~50 for BERT-based classifiers, 15~–~30 for sequence-to-sequence models, and 5~–~20 for fine-tuned LLMs. These ranges are proportionally reduced for zero-resource setups that rely on larger pseudo-training sets (e.g., cross-lingual transfer or code-switching) to prevent overfitting. Second, models are retrained on the complete training data using the best-performing epoch count and evaluated on the held-out test set. Each configuration is repeated with five random seeds, and final results are averaged to mitigate stochastic variation. By pairing a multilingual baseline with a monolingual upper bound, we derive both a realistic transfer estimate and a ceiling for in-language performance.

We test significance (\textit{p}\textsubscript{adj} $\leq$ 0.05) using parametric or non-parametric tests (ANOVA/\textit{t}-test or Friedman/Wilcoxon ~\cite{Field2012-ul}) with Bonferroni-Holm correction~\cite{Holm1979-zm} to assess differences in performance between multilingual and language-specific models, language-specific variability in training outcomes, and variation in zero-resource transfer effectiveness across tasks and languages.

\subsubsection{In-Language Supervised Settings}



To ensure fair cross-lingual comparability and consistent label distributions across languages, we employ a balanced dataset configuration. This setup equalizes dataset sizes and aspect category distributions as described in Section~\ref{sec:preprocessing}. Multilingual models are fine-tuned to simulate scenarios where moderately sized target-language data are available but no language-specific model exists, while monolingual pretrained models serve as upper bounds for in-language performance.

\subsubsection{Zero-Resource Settings}

To evaluate the transfer capabilities of supervised multilingual models in the absence of target-language training data, we utilize three zero-resource ABSA strategies:
\begin{itemize}
    \item \textbf{Cross-Lingual Transfer (CLT):} The model is trained on all balanced training sets except for the target language.
    \item \textbf{Code-Switching (CS):} Augments each English training sentence with several variants, original, translated, and mixed, by replacing key lexical items (e.g., aspect and sentiment terms) with their target-language counterparts using LLMs. This hybrid data provides weak cross-lingual supervision and exposes the model to bilingual lexical patterns while maintaining grammatical structure.
    \item \textbf{Machine Translation (MT):} The English training data is automatically translated into the target language using LLMs.
\end{itemize}

To strengthen the generalizability of our work and based on the assumption that optimized monolingual models are generally unavailable under such conditions, all zero-resource experiments are conducted using multilingual models only.

%% file: sections/results.tex
\subsection{Results for Monolingual Training}

\vspace{-0.5em}

In the monolingual setting, where training and testing use the same language, we compare two configurations on the balanced datasets (see Table~\ref{tab:results-balanced}): Multi = a multilingual model (e.g., \textit{mT5-base}) fine-tuned per language, and Spec = a language-specific model (e.g., \textit{ruT5-base} for Russian) fine-tuned on the same data. This allows a direct comparison between multilingual and monolingual pretraining. We make four main observations:

\vspace{-0.5em}

\paragraph{\textbf{Inter‐language variability}}
Performance under the \textbf{Multi} configuration varies considerably across languages and tasks. In both ACD and ACSA, multilingual models achieve their strongest results for German and Spanish, while a language such as French shows consistently weaker performance. The same trend appears in TASD, where the disparity between high- and low-performing languages widens further. However, these cross-linguistic differences are only statistically significant in a few isolated cases. Interestingly, this pattern holds across architectures: encoder-only classifiers, graph-based models, and fine-tuned LLMs show similar cross-lingual ranking orders, suggesting that the limitations stem less from model design and more from the representational mismatch between multilingual embeddings and the typological diversity of target languages.

\begin{table}[ht]
\centering
\fontsize{8}{9.6}\selectfont
\renewcommand{\arraystretch}{1.0}
\setlength{\tabcolsep}{2pt}

\begin{subtable}{\columnwidth}
\centering
\caption{ACD}
\resizebox{\columnwidth}{!}{
\begin{tabular}{lp{1.2cm}ccccccc}
\toprule
\textbf{Method} & \textbf{Setting} & \textbf{CS} & \textbf{DE} & \textbf{EN} & \textbf{ES} & \textbf{FR} & \textbf{NL} & \textbf{RU} \\
\midrule
\multirow{2}{*}{BERT-CLF}
& Multi  & 77.18 & 78.19 & 75.42 & 76.62 & 71.48 & 73.69 & 76.89  \\
& Spec & 80.98 & 82.56 & 80.79 & 79.53 & 71.81 & 75.62 & 81.16  \\
\midrule
\multirow{2}{*}{\makecell[l]{LLaMA 3.1\\8B FT}}
& \multirow{2}{*}{\makecell[l]{Multi}} & \multirow{2}{*}{\makecell[l]{75.64}} & \multirow{2}{*}{\makecell[l]{78.40}} & \multirow{2}{*}{\makecell[l]{83.82}} & \multirow{2}{*}{\makecell[l]{80.33}} & \multirow{2}{*}{\makecell[l]{76.84}} & \multirow{2}{*}{\makecell[l]{80.27}} & \multirow{2}{*}{\makecell[l]{79.40}}  \\\\
\midrule
\multirow{2}{*}{\makecell[l]{Gemma 3\\27B 50-shot}}
  & \multirow{2}{*}{\makecell[l]{50-shot}}    & \multirow{2}{*}{\makecell[l]{74.52}} & \multirow{2}{*}{\makecell[l]{77.38}} & \multirow{2}{*}{\makecell[l]{77.38}} & \multirow{2}{*}{\makecell[l]{75.16}} & \multirow{2}{*}{\makecell[l]{70.06}} & \multirow{2}{*}{\makecell[l]{78.63}} & \multirow{2}{*}{\makecell[l]{75.23}} \\\\
\bottomrule
\end{tabular}
}

\end{subtable}

\vspace{0.2em}

\begin{subtable}{\columnwidth}
\centering
\caption{ACSA}
\resizebox{\columnwidth}{!}{
\begin{tabular}{lp{1.2cm}ccccccc}
\toprule
\textbf{Method} & \textbf{Setting} & \textbf{CS} & \textbf{DE} & \textbf{EN} & \textbf{ES} & \textbf{FR} & \textbf{NL} & \textbf{RU} \\
\midrule
\multirow{2}{*}{BERT-CLF}
& Multi & 61.37 & 64.73 & 57.15 & 63.92 & 52.37 & 54.24 & 57.56  \\
& Spec & 69.63 & 72.56 & 65.34 & 68.91 & 50.56 & 61.48 & 63.29  \\
\midrule
\multirow{2}{*}{Hier-GCN}
& Multi & 65.39 & 67.74 & 63.68 & 68.25 & 58.39 & 60.36 & 61.64  \\
& Spec & 70.66 & 76.09 & 70.22 & 72.46 & 60.97 & 67.11 & 67.56  \\
\midrule
\multirow{1}{*}{\makecell[l]{LLaMA 3.1\\8B FT}}
& \multirow{2}{*}{\makecell[l]{Multi}} & \multirow{2}{*}{\makecell[l]{70.66}} & \multirow{2}{*}{\makecell[l]{76.30}} & \multirow{2}{*}{\makecell[l]{79.89}} & \multirow{2}{*}{\makecell[l]{76.18}} & \multirow{2}{*}{\makecell[l]{70.02}} & \multirow{2}{*}{\makecell[l]{74.07}} & \multirow{2}{*}{\makecell[l]{73.00}}  \\\\
\midrule
\multirow{1}{*}{\makecell[l]{Gemma 3\\27B}}
  & \multirow{2}{*}{\makecell[l]{50-shot}}    & \multirow{2}{*}{\makecell[l]{70.12}} & \multirow{2}{*}{\makecell[l]{76.51}} & \multirow{2}{*}{\makecell[l]{75.68}} & \multirow{2}{*}{\makecell[l]{72.49}} & \multirow{2}{*}{\makecell[l]{66.78}} & \multirow{2}{*}{\makecell[l]{75.06}} & \multirow{2}{*}{\makecell[l]{73.68}} \\\\
\bottomrule
\end{tabular}
}
\end{subtable}

\vspace{0.2em}

\begin{subtable}{\columnwidth}
\centering
\caption{TASD}
\resizebox{\columnwidth}{!}{
\begin{tabular}{lp{1.2cm}ccccccc}
\toprule
\textbf{Method} & \textbf{Setting} & \textbf{CS} & \textbf{DE} & \textbf{EN} & \textbf{ES} & \textbf{FR} & \textbf{NL} & \textbf{RU} \\
\midrule
\multirow{2}{*}{DLO}
& Multi & 54.60 & 48.10 & 50.37 & 57.94 & 46.70 & 46.35 & 49.68  \\
& Spec & -- & 59.59 & 70.55 & 58.74 & 50.99 & 59.12 & 59.87  \\
\midrule
\multirow{1}{*}{\makecell[l]{LLaMA 3.1\\8B FT}}
& \multirow{2}{*}{\makecell[l]{Multi}} & \multirow{2}{*}{\makecell[l]{61.09}} & \multirow{2}{*}{\makecell[l]{63.27}} & \multirow{2}{*}{\makecell[l]{69.95}} & \multirow{2}{*}{\makecell[l]{64.09}} & \multirow{2}{*}{\makecell[l]{56.16}} & \multirow{2}{*}{\makecell[l]{62.16}} & \multirow{2}{*}{\makecell[l]{62.40}}  \\\\
\midrule
\multirow{1}{*}{\makecell[l]{Gemma 3\\27B}}
  & \multirow{2}{*}{\makecell[l]{50-shot}}    & \multirow{2}{*}{\makecell[l]{60.58}} & \multirow{2}{*}{\makecell[l]{62.94}} & \multirow{2}{*}{\makecell[l]{68.53}} & \multirow{2}{*}{\makecell[l]{59.29}} & \multirow{2}{*}{\makecell[l]{55.14}} & \multirow{2}{*}{\makecell[l]{58.40}} & \multirow{2}{*}{\makecell[l]{58.90}} \\\\
\bottomrule
\end{tabular}
}
\end{subtable}

\caption{Supervised F1-Micro scores per language for each method and dataset/model resource setting, split by task. Abbr.: Spec = language-specific model; Multi = multilingual model. For TASD, Czech results are omitted (“--”) as no language-specific T5 model exists.}
\vspace{-2em}
\label{tab:results-balanced}
\end{table}

\vspace{-0.5em}

\paragraph{\textbf{Consistent monolingual gains when using language-specific models}}
Across all languages and modeling paradigms, using language-specific pretrained models (\textbf{Spec}) consistently improves performance over their multilingual counterparts (\textbf{Multi}). The improvement is modest for simpler classification tasks such as ACD ($\approx$~3.3 F1 points on average) and ACSA ($\approx$~5.6 points), but becomes substantially larger for the more complex TASD task ($\approx$~8.5 points). These gains are statistically significant across all tasks, confirming the robustness of the monolingual advantage and consistent performance benefit that multilingual architectures have not yet fully bridged. Even advanced LLMs still have weaknesses in certain languages and tasks, suggesting that monolingual pretraining remains a key factor for achieving SOTA ABSA performance. Similar trends are observed in other NLP tasks, where monolingual models, such as FinBERT \cite{virtanen2019multilingualenoughbertfinnish}, CamemBERT \cite{Martin2020-ph}, RobeCzech \cite{Straka2021-zu}, and others \cite{Ulcar2026-ja}, consistently outperform multilingual counterparts across multiple languages and benchmarks.

\vspace{-0.5em}

\paragraph{\textbf{Superior performance of fine‐tuned LLMs}}
The fine-tuned LLaMA 3.1 8B achieves the strongest overall results in our supervised comparisons, consistently outperforming both multilingual and monolingual encoder-based models across most languages and tasks. Its advantage is most pronounced for complex generative subtasks like TASD, confirming the benefit of large-scale decoder architectures for structured sentiment extraction. However, this superiority is not universal. In some languages and simpler tasks (e.g., ACD or ACSA), fine-tuned BERT-based classifiers still match or even exceed multilingual LLM performance. These findings highlight a nuanced picture: fine-tuned LLMs excel in complex, structured ABSA subtasks and offer the highest performance for multilingual adaptation, but they do not yet replace specialized models as a universal solution across all languages and task granularities.

Notably, the few-shot Gemma 3 27B model shows strong robustness without fine-tuning, surpassing supervised baselines in ACSA for German, Dutch, and Russian, and performing on par with most language-specific SOTA models elsewhere. While it falls short in ACD and TASD, its consistent performance underscores the potential of few-shot prompting as a lightweight alternative to full fine-tuning.


\vspace{-0.5em}

\paragraph{First insights into multilingual ASQP experiments}
For the most complex ASQP task (available only in English and German), the overall impression remains (see Table \ref{tab:cross-lingual-results-asqp}): both supervised methods, the generative DLO and the fine-tuned LLaMA 3.1 8B, achieve the highest scores (59.35 / 48.16 F1-micro for DLO vs. 57.93 / 53.44 F1-micro for LLaMA). In contrast, the few-shot Gemma 3 27B trails behind (51.10 / 41.47 F1-micro), despite remaining well above zero-shot levels. These results underline that the performance gap between few-shot prompting and fully supervised learning widens as task complexity increases.

\begin{table}[!ht]
\centering
\fontsize{8}{9.6}\selectfont
\renewcommand{\arraystretch}{1.0}
\setlength{\tabcolsep}{2pt}


\label{tab:restructured-results}

\vspace{-0.5em}

\begin{subtable}{\columnwidth}
\centering
\caption{ACD}
\resizebox{\columnwidth}{!}{
\begin{tabular}{llcccccc}
\toprule
\textbf{Method} & \textbf{Setting} & \textbf{CS} & \textbf{DE} & \textbf{ES} & \textbf{FR} & \textbf{NL} & \textbf{RU} \\
\midrule
\multirow{4}{*}{BERT-CLF}
  & Supervised & 77.18 & 78.19 & 76.62 & 71.48 & 73.69 & 76.89  \\
    \cmidrule{2-8}
  & CLT    & 61.51 & 71.83 & 67.46 & 67.74 & 64.86 & 67.48  \\
  & CS         & 69.05 & 71.41 & 69.42 & 68.36 & 67.31 & 69.76  \\
  & MT         & 66.82 & 70.21 & 67.66 & 66.34 & 66.02 & 67.81  \\
\midrule
\multirow{4}{*}{\makecell[l]{LLaMA 3.1\\8B FT}}
  & Supervised  & 83.31 & 84.52 & 84.80 & 81.65 & 85.56 & 87.28 \\
    \cmidrule{2-8}
  & CLT     & \textbf{82.27} & \textbf{84.51} & \textbf{81.90} & \textbf{81.49} & \textbf{83.64} & \textbf{86.45} \\
  & CS         & 80.02 & 81.43 & 81.39 & 80.22 & 80.02 & 84.76 \\
  & MT         & 78.26 & 80.49 & 80.66 & 79.84 & 80.06 & 83.71 \\
\midrule
\multirow{2}{*}{\makecell[l]{Gemma 3\\27B}}
  & 50-shot    & 74.52 & 77.38 & 75.16 & 70.06 & 78.63 & 75.23 \\
    \cmidrule{2-8}
  & 0-shot     & 66.11 & 68.42 & 66.35 & 59.10 & 69.50 & 73.28 \\
\bottomrule
\end{tabular}
}
\end{subtable}

\vspace{-0.3em}
 
\begin{subtable}{\columnwidth}
\centering
\caption{ACSA}
\resizebox{\columnwidth}{!}{
\begin{tabular}{llcccccc}
\toprule
\textbf{Method} & \textbf{Setting} & \textbf{CS} & \textbf{DE} & \textbf{ES} & \textbf{FR} & \textbf{NL} & \textbf{RU} \\
\midrule
\multirow{4}{*}{BERT-CLF}
  & Supervised & 61.37 & 64.73 & 63.92 & 52.37 & 54.24 & 57.56  \\
    \cmidrule{2-8}
  & CLT    & 40.57 & 50.78 & 54.32 & 43.80 & 46.98 & 49.18  \\
  & CS         & 48.21 & 55.78 & 50.10 & 48.92 & 48.98 & 53.59  \\
  & MT         & 44.94 & 53.50 & 48.95 & 45.74 & 47.68 & 49.41  \\
\midrule
\multirow{4}{*}{Hier-GCN}
  & Supervised & 65.39 & 67.74 & 68.25 & 58.39 & 60.36 & 61.64  \\
  \cmidrule{2-8}
  & CLT    & 42.50 & 52.50 & 56.16 & 44.96 & 45.39 & 44.89  \\
  & CS         & 55.78 & 59.67 & 60.49 & 50.21 & 55.99 & 56.33  \\
  & MT         & 54.40 & 58.72 & 59.48 & 48.11 & 53.98 & 55.23  \\
\midrule
\multirow{4}{*}{\makecell[l]{LLaMA 3.1\\8B FT}}
  & Supervised  & 75.16 & 81.48 & 79.40 & 73.30 & 78.99 & 79.50 \\
    \cmidrule{2-8}
  & CLT     & \textbf{73.76} & \textbf{80.75} & \textbf{77.40} & \textbf{73.41} & \textbf{77.90} & \textbf{78.46} \\
  & CS          & 71.42 & 77.41 & 76.20 & 70.36 & 74.63 & 75.51 \\
  & MT          & 70.03 & 76.61 & 74.38 & 70.64 & 73.86 & 74.96 \\
\midrule
\multirow{2}{*}{\makecell[l]{Gemma 3\\27B}}
  & 50-shot    & 70.12 & 76.51 & 72.49 & 66.78 & 75.06 & 73.68 \\
    \cmidrule{2-8}
  & 0-shot     & 69.39 & 73.50 & 68.21 & 62.40 & 70.27 & 68.05 \\
\bottomrule
\end{tabular}
}

\end{subtable}

\vspace{-0.3em}

\begin{subtable}{\columnwidth}
\centering
\caption{TASD}
\resizebox{\columnwidth}{!}{
\begin{tabular}{llcccccc}
\toprule
\textbf{Method} & \textbf{Setting} & \textbf{CS} & \textbf{DE} & \textbf{ES} & \textbf{FR} & \textbf{NL} & \textbf{RU} \\
\midrule
\multirow{4}{*}{DLO}
  & Supervised & 54.60 & 48.10 & 57.94 & 46.70 & 46.35 & 49.68  \\
  \cmidrule{2-8}
  & CLT    & 24.74 & 46.51 & 38.87 & 38.08 & 38.60 & 18.51  \\
  & CS         & 43.30 & 48.73 & 51.37 & 39.12 & 38.10 & 43.94  \\
  & MT         & 40.28 & 39.86 & 48.90 & 36.71 & 40.49 & 43.70  \\
\midrule
\multirow{4}{*}{\makecell[l]{LLaMA 3.1\\8B FT}}
  & Supervised  & 65.43 & 67.89 & 68.30 & 61.98 & 65.85 & 65.72 \\
  \cmidrule{2-8}
  & CLT     & 53.54 & \textbf{66.28} & 60.18 & \textbf{58.12} & \textbf{62.98} & 57.61 \\
  & CS          & \textbf{56.86} & 62.86 & \textbf{61.89} & 55.99 & 59.27 & \textbf{59.26} \\
  & MT          & 55.27 & 61.21 & 60.73 & 53.99 & 59.08 & 57.75 \\
\midrule
\multirow{2}{*}{\makecell[l]{Gemma 3\\27B}}
  & 50-shot    & 60.58 & 62.94 & 59.29 & 55.14 & 58.40 & 58.90 \\
  \cmidrule{2-8}
  & 0-shot     & 47.74 & 47.95 & 40.43 & 34.53 & 36.22 & 38.03 \\
\bottomrule
\end{tabular}
}

\end{subtable}

\vspace{-0.3em}

\begin{subtable}{\columnwidth}
\centering
\caption{ASQP}
\begin{tabular}{llcc}
\toprule
\textbf{Method} & \textbf{Setting} & \textbf{EN} & \textbf{DE} \\
\midrule
\multirow{3}{*}{DLO}
  & Supervised & 59.35 & 48.16  \\
  \cmidrule{2-4}
  & CS         & -- & \textbf{36.92}  \\
  & MT         & -- & 36.24  \\
\midrule
\multirow{3}{*}{\makecell[l]{LLaMA 3.1\\8B FT}}
  & Supervised  & 57.93 & 53.44 \\
  \cmidrule{2-4}
  & CS          & -- & 41.06 \\
  & MT          & -- & \textbf{42.95} \\
\midrule
\multirow{2}{*}{\makecell[l]{Gemma 3\\27B}}
  & 50-shot    & 51.10 & 41.47 \\
  \cmidrule{2-4}
  & 0-shot     & 28.96 & 17.90 \\
\bottomrule
\end{tabular}
\label{tab:cross-lingual-results-asqp}

\end{subtable}
\caption{F1-micro scores per language and method under zero-resource conditions. CLT: trained on all languages except the target; CS: code-switched data; MT: machine-translated data. Supervised/50-shot scores are included for comparison. Bold values indicate the best-performing configuration.}
\label{tab:cross-lingual-results}
\vspace{-4em}
\end{table}

\subsection{Results for Multi-/Crosslingual Training}
After establishing the supervised monolingual upper bound, we now assess ABSA portability under multi- and cross-lingual training, examining generalization from non-target languages or augmented data and how far multilingual pretraining substitutes in-language supervision. We compare the effects of task complexity, augmentation strategy, and model architecture on cross-lingual transfer.

\vspace{-0.5em}

\paragraph{Overall cross-lingual transfer performance and task complexity effects.}
Across all tasks and languages, a clear degradation is observed when moving from supervised to zero-resource conditions. Models trained on all non-target languages (CLT) consistently underperform their supervised counterparts, confirming the persistent gap between fine-tuning and actual transfer generalization. The magnitude of this gap, however, varies strongly with task complexity: for simpler classification tasks such as ACD and ACSA, multilingual fine-tuned models still provide competitive performance, often within 5~–~8 F1 points of the supervised baseline. In contrast, for more structurally complex tasks like TASD and ASQP, transfer performance drops substantially, with relative declines exceeding 20 points in several cases. This pattern underscores that while multilingual fine-tuning is sufficient for coarse-grained sentiment classification, the extraction of structured opinion relations (e.g., TASD triplets) still benefits from task- or language-specific adaptation. Nevertheless, all multilingual models achieve actionable results even without language-specific pretraining, enabling meaningful ABSA evaluation for low-resource or underrepresented languages where dedicated models and datasets are unavailable. Notably, the fine-tuned LLaMA 3.1 8B model demonstrates the most robust cross-lingual generalization, maintaining over 81 F1-micro in ACD and roughly 73~–~80 F1-micro in ACSA even under zero-resource conditions, whereas encoder-only architectures (BERT, Hier-GCN) suffer the largest degradation.


\vspace{-0.5em}

\paragraph{Impact of zero-resource strategies.}

We compare three zero-resource transfer strategies, CLT, CS, and MT, relative to supervised upper bounds. For encoder-based and seq-to-seq models, CS yields the strongest relative performance, exceeding CLT and MT by 3–6 F1 points on average and reducing the gap to supervised baselines. 
This supports findings by \citet{Zhang2021-kz} and \citet{Wu2025-hq}, who attribute CS gains to increased lexical diversity, while \citet{Smid2025-kw} and \citet{Zhang2025-ih} show that adding synonym replacement or distillation further stabilizes transfer, while outperforming direct translation-based approaches.
In contrast, MT-based augmentation brings smaller improvements, suggesting that data quantity and variation matter more than grammatical accuracy, particularly for encoder-based models, which appear to benefit from the higher lexical diversity introduced by code-switching. 
Fine-tuned LLMs usually achieve their best results under CLT conditions, often matching or approaching supervised performance, indicating that large decoder-based models already internalize cross-lingual alignment without additional augmentation. 
Overall, while CS and MT are valuable for smaller architectures, CLT proves most effective for large fine-tuned LLMs in zero-resource evaluation.


\vspace{-0.5em}

\paragraph{Comparative robustness of fine-tuned vs. few-shot LLMs.}
The fine-tuned LLaMA 3.1 8B consistently achieves the highest zero-resource transfer scores across all subtasks, with at least one zero-resource configuration performing significantly better than all other approaches.
However, the few-shot Gemma 3 27B model exhibits remarkable stability across languages without any fine-tuning: for ACD and ACSA, it almost matches the supervised encoder baselines, while for more complex tasks such as TASD and ASQP, the performance drop is more pronounced.
This cross-lingual consistency underscores the utility of instruction models for low-resource scenarios where no language-specific supervision (neither datasets nor specialized models) is available.
These observations align with the findings of \citet{Smid2025-xz}, who show that LLM-based augmentation and in-context learning can approach supervised performance in low-resource settings, though performance gains diminish for larger model scales.

%% file: sections/conclusion.tex
This work presented a comprehensive multilingual evaluation of SOTA approaches for ABSA across seven languages and four subtasks. By comparing encoder-only, sequence-to-sequence, and decoder-only architectures under varying resource conditions, we analyzed how well current models generalize across languages and ABSA tasks.

Our results reveal a clear hierarchy: instruction fine-tuned LLMs achieve the highest overall scores, particularly in complex generative tasks such as TASD and ASQP. Smaller encoder-based models remain competitive for simpler classification tasks (ACD, ACSA), offering strong performance with lower computational costs.


Language-specific models still outperform multilingual ones, though this gap narrows with larger, more multilingual architectures. Code-switching yields the most consistent improvements in zero-resource settings, while cross-lingual training on non-target languages allows fine-tuned LLMs to approach supervised performance. Gemma 3 27B achieves competitive zero-shot results on simpler tasks but declines on more complex ones such as ASQP; in few-shot mode, however, it stabilizes and approaches fine-tuned performance, making it a practical low-resource alternative.

Beyond empirical findings, we contribute two new German resources: an adapted \textit{GERestaurant} dataset aligned with the SemEval aspect-category schema, and the first German ASQP dataset (\textit{GERest}) for structured opinion extraction. These additions extend ABSA research beyond English and enable controlled cross-lingual ASQP evaluation.

Future research should extend multilingual ABSA evaluation to additional domains and typologically diverse languages to test the generalizability of current methods. Furthermore, while we cover representative SOTA models across major paradigms, additional approaches, such as hybrid syntactic LLM approaches \cite{Negi2024-kt} or dual-stream data synthesis frameworks \cite{Xu2025-kk}, which combine structural linguistic information or synthetic data generation, could provide further insights. Finally, evaluating instruction-tuned LLMs in few-shot and semi-supervised scenarios across new domains will be key to understanding their practical potential for multilingual sentiment analysis at scale.

%% file: sections/limitations.tex
While this study provides a broad multilingual benchmark for ABSA, several limitations remain. As the benchmark datasets used here were released before the pretraining of the evaluated models, potential data contamination cannot be entirely ruled out, even though no direct overlap between models and datasets is documented. Nevertheless, these resources represent the de facto standard for multilingual and cross-lingual ABSA research and were therefore used to ensure comparability with prior work. 

In addition, to ensure comparable label distributions across languages and experimental settings, we employed balanced versions of the datasets, which may differ from naturally occurring sentiment and aspect distributions in the original data and therefore are less representative of real-world scenarios. Moreover, only a few datasets offer a sufficiently broad and multilingual foundation to enable systematic cross-lingual evaluation at this scale. 

To further validate and generalize our findings, future studies should extend cross-lingual ABSA to additional domains (e.g., product reviews or social media), such as the English OATS~\cite{Chebolu2024-pz} or FlightABSA~\cite{Hellwig2025-sd} datasets. 

Additionally, although we include seven languages, further evaluations on typologically diverse or low-resource languages are needed to better assess transfer robustness. 

Moreover, while our experiments cover representative encoder-, sequence-to-seqence-, and decoder-based architectures, more recent transformer variants such as ModernBERT~\cite{warner2024smarterbetterfasterlonger} could provide additional insights into cross-lingual generalization. We deliberately refrained from including these architectures in this study, as language-specific pretrained versions of such models (e.g., ModernGBERT \cite{ehrmanntraut2025moderngbertgermanonly1bencoder}) are still scarce, limiting their comparability to existing baselines.

From an ethical standpoint, no new user data were collected for this study. The newly contributed German datasets (adapted \textit{GERestaurant} and new \textit{GERest}) were derived from existing, anonymized corpora and contain no personally identifiable information. We used Claude~4.0\footnote{Claude Sonnet: \url{https://www.anthropic.com/claude/sonnet}} for support in code optimization and linguistic editing; all methodological decisions, analyses, and reported results were manually created and verified by the authors to avoid automated bias or factual distortion.

%% file: sections/appendix.tex
\appendix
\onecolumn

\section{GERest}
\label{appendix:gerest}

GERest is derived from the TASD dataset GERestaurant, introduced by \citet{Hellwig2024-wi}. This dataset was prepared with the aim of mirroring the structure of ASQP-Rest16, ensuring comparable quantities of training, validation, and test examples. The original GERestaurant comprises a training set with 2,154 examples and a test set with 924 examples, but does not include a dedicated validation set. To address this issue, a subset of the training examples was allocated as a validation set for GERest. All examples were refined by introducing an additional opinion term. The final dataset distribution is as follows:

\begin{itemize}
  \item \textbf{Training:} 1,264 examples derived from GERestaurant's training set.
  \item \textbf{Validation:} 316 examples derived from GERestaurant's training set.
  \item \textbf{Test:} 544 examples derived from GERestaurant's test set.
\end{itemize}

Moreover, the 13 aspect categories used in the ASQP dataset by \citet{Zhang2021-qu} were adopted for GERest. Annotators revised the examples from GERestaurant to comprise quadruples including one of the 13 aspect categories instead of the five aspect categories considered for GERestaurant.


The annotation process for GERest followed the ASQP annotation guidelines established by \citet{Zhang2021-ga} and \citet{wan2020target} for Rest15 and Rest16. 

All label revisions were initially performed by a computer science bachelor’s student (Annotator \textit{A}). Subsequently, all examples were reviewed and refined by a PhD student (Annotator \textit{B}) with prior experience in annotating ABSA datasets.

Among the 2,124 annotated sentences, annotator \textit{B} proposed an alternative label to that proposed by annotator \textit{A} in the case of 184 sentences. Of these 184 proposed changes, 179 were accepted by annotator \textit{A}. For the remaining 5 cases, a joint decision was made: in 3 instances, the original annotation by annotator \textit{A} was retained, while in 2 cases, annotator \textit{B}'s label was adopted.

\section{Prompt Examples for English}\label{app:prompt-examples}

\begin{figure}[h]
    \centering
    {
    \setlength{\fboxsep}{5pt}
    \setlength{\fboxrule}{0.5pt}
    \fbox{\includegraphics[width=1\linewidth]{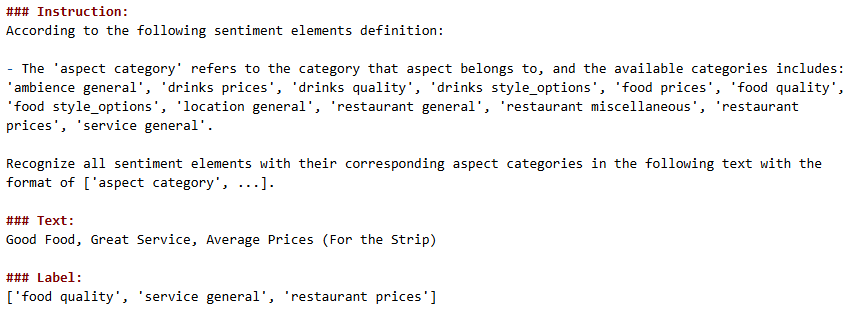}}
    }
    \caption{Prompt example for the ACD task for the English-language SemEval 2016 restaurant dataset.}
    \label{app:prompt-example-acd}
\end{figure}

\begin{figure*}[h]
    \centering
    {
    \setlength{\fboxsep}{5pt}
    \setlength{\fboxrule}{0.5pt}
    \fbox{\includegraphics[width=1\linewidth]{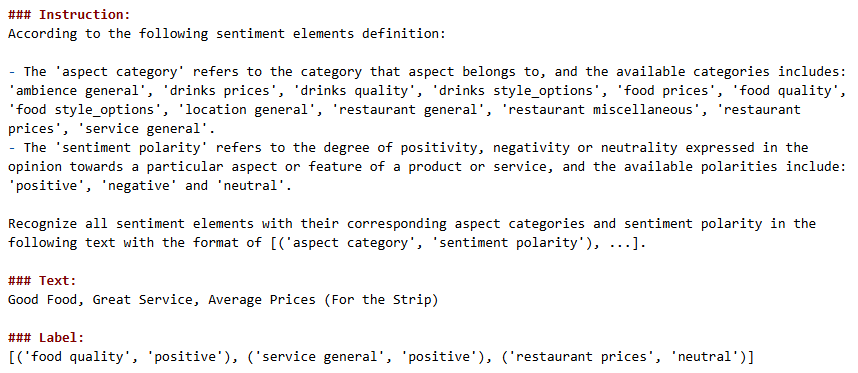}}
    }
    \caption{Prompt example for the ACSA task for the English-language SemEval 2016 restaurant dataset.}
    \label{app:prompt-example-acsa}
\end{figure*}

\begin{figure*}[h]
    \centering
    {
    \setlength{\fboxsep}{5pt}
    \setlength{\fboxrule}{0.5pt}
    \fbox{\includegraphics[width=1\linewidth]{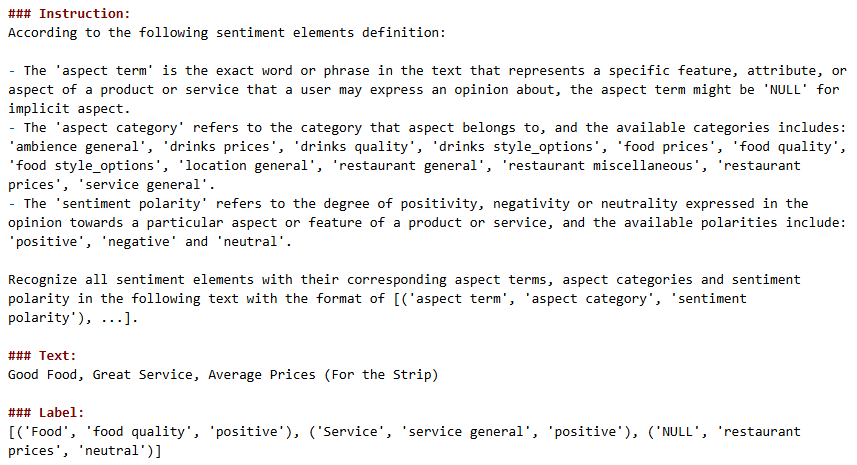}}
    }
    \caption{Prompt example for the TASD task for the English-language SemEval 2016 restaurant dataset.}
    \label{app:prompt-example-tasd}
\end{figure*}